\pdfoutput=1

\documentclass[11pt]{article}

\usepackage{amsmath, amsfonts, amssymb, amsthm, xspace, color}
\usepackage{booktabs} 

\usepackage{subfigure, multirow, tabularx} 
\usepackage{booktabs} 
\usepackage{enumitem}
\usepackage{flushend}
\usepackage[T1]{fontenc}
\usepackage{epstopdf}
\usepackage{balance}
\usepackage{graphicx}
\usepackage[noend,ruled,linesnumbered]{algorithm2e} 
\usepackage{url}
\usepackage{wrapfig,lipsum}
\usepackage{makecell}
\usepackage{wrapfig}

\newcommand{\hide}[1]{} 

\newcommand{\eg}{e.g.\xspace} 
\newcommand{\nop}[1]{}
\newcommand{\mquote}[1]{{``\emph{#1}''}}

\newtheorem{thm:def}{Definition}
\newtheorem{thm:eg}{Example}
\newtheorem{thm:lem}{Lemma}
\newtheorem{thm:obs}{Observation}
\newtheorem{thm:req}{Requirement}
\newtheorem{thm:prop}{Proposition}
\newtheorem{thm:principle}{Principle}
\newtheorem{thm:thm}{Theorem}
\newtheorem{thm:corollary}{Corollary}

\newcommand{\pair}[1]{$\langle$#1$\rangle$}			

\DeclareMathOperator*{\argmax}{arg\,max}

\newcommand{\ours}{\mbox{\sc ETypeClus}\xspace}

\usepackage{bbm}
\usepackage{adjustbox}
\usepackage[super]{nth}

\usepackage[]{emnlp2021}

\usepackage{times}
\usepackage{latexsym}

\usepackage[T1]{fontenc}

\usepackage[utf8]{inputenc}

\usepackage{microtype}

\definecolor{OrangeRed}{rgb}{1.0, 0.27, 0.0}
\definecolor{midnightgreen}{rgb}{0.0, 0.29, 0.33}

\title{Corpus-based Open-Domain Event Type Induction}
\author{Jiaming Shen, Yunyi Zhang, Heng Ji, Jiawei Han\\
\small Department of Computer Science, University of Illinois Urbana-Champaign, IL, USA \\
\footnotesize \{js2, yzhan238, hengji, hanj\}@illinois.edu \\
}

\begin{document}
\maketitle
\begin{abstract}

Traditional event extraction methods require predefined event types and their corresponding annotations to learn event extractors. 
These prerequisites are often hard to be satisfied in real-world applications.
This work presents a corpus-based open-domain event type induction method that automatically discovers a set of event types from a given corpus. 
As events of the same type could be expressed in multiple ways, we propose to represent each event type as a cluster of $\langle$predicate sense, object head$\rangle$ pairs.
Specifically, our method (1) selects salient predicates and object heads, (2) disambiguates predicate senses using only a verb sense dictionary, and (3) obtains event types by jointly embedding and clustering $\langle$predicate sense, object head$\rangle$ pairs in a latent spherical space. 
Our experiments, on three datasets from different domains, show our method can discover salient and high-quality event types, according to both automatic and human evaluations\footnote{\small The programs, data and resources are publicly available for research purpose at \url{https://github.com/mickeystroller/ETypeClus}.}.

\end{abstract}


\section{Introduction}\label{sec:intro}
One step towards converting massive unstructured text into structured, machine-readable representations is event extraction---the identification and typing of event triggers and arguments in text. 
Most event extraction methods~\cite{Ahn2006TheSO,Ji2008RefiningEE,Du2020EventEB,Li2021DocumentLevelEA} assume a set of predefined event types and their corresponding annotations are curated by human experts.
This annotation process is expensive and time-consuming. 
Besides, those manually-defined event types often fail to generalize to new domains. 
For example, the widely used ACE 2005 event schemas\footnote{\small \url{https://www.ldc.upenn.edu/collaborations/past-projects/ace}} do not contain any event type about \texttt{Transmit Virus} or \texttt{Treat Disease} and thus cannot be readily applied to extract pandemic events. 

\begin{figure}[!t]
	\centering
	\centerline{\includegraphics[width=0.47\textwidth]{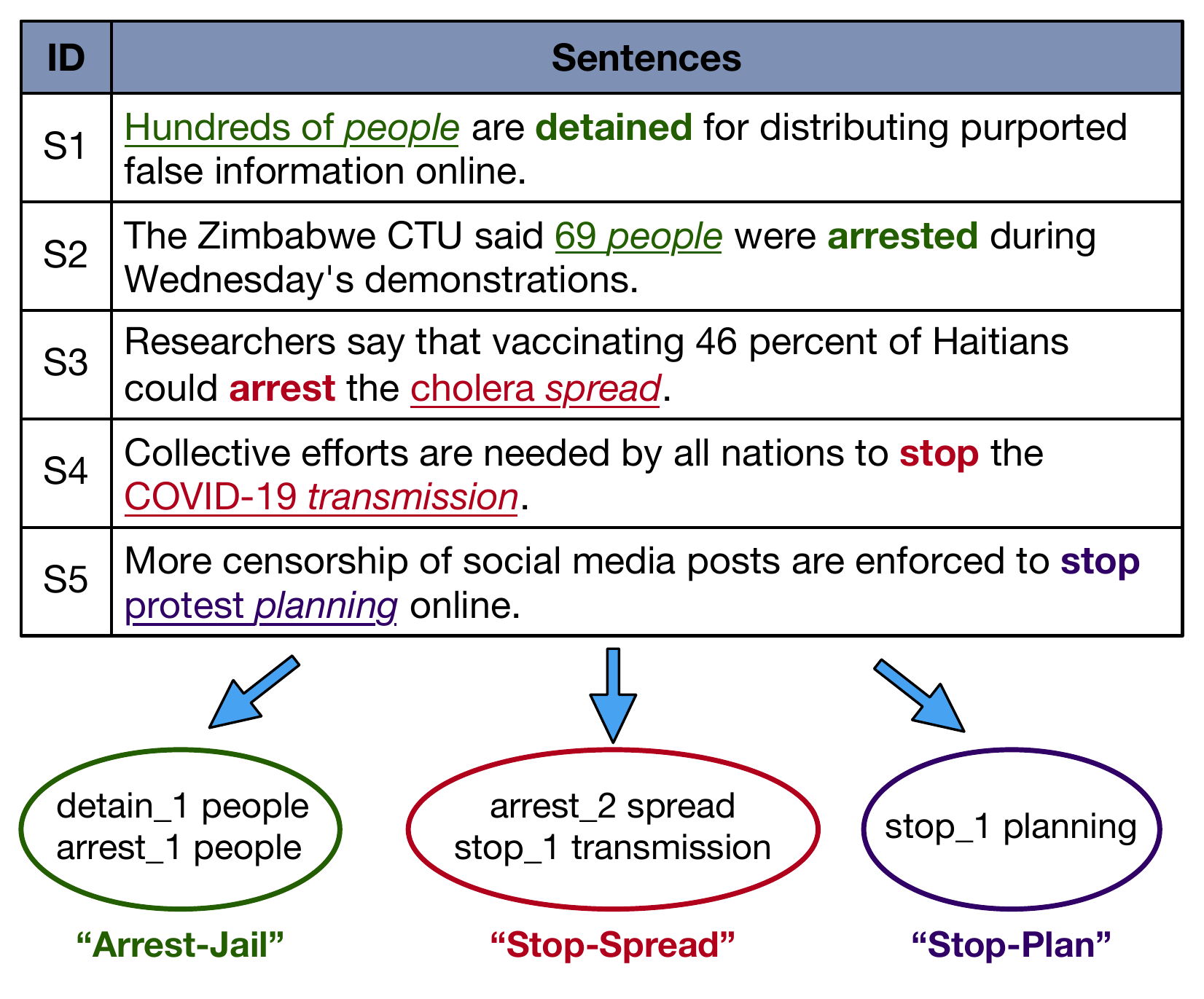}}
	\vspace{-0.2cm}
	\caption{Motivating example sentences and induced event types. \textbf{Predicates} are in bold. \underline{Objects} are underlined and \emph{object heads} are in italics. Colors indicate event types. The suffix number followed by each predicate verb lemma indicates the predicate verb sense.}
	\label{fig:intro-example}
	\vspace{-0.3cm}
\end{figure}

To automatically induce event schemas from raw text, researchers have studied ad-hoc clustering-based algorithms~\cite{Sekine2006OnDemandIE,Chambers2011TemplateBasedIE} and probabilistic generative methods~\cite{Chambers2013EventSI,Cheung2013ProbabilisticFI,Nguyen2015GenerativeES} to discover a set of event types and argument roles.
These methods typically utilize bag-of-word text representations and impose strong statistical assumptions.
\citet{Huang2016LiberalEE} relax those restrictions using a pipelined approach that leverages extensive lexical and semantic resources (\eg, FrameNet~\cite{Baker1998TheBF}, VerbNet~\cite{Schuler2005VerbnetAB}, and PropBank~\cite{Palmer2005ThePB}) to discover event schemas.
While being effective, this method is limited by the scope of external resources and accuracies of its preprocessing tools.
Recently, some studies~\cite{Huang2018ZeroShotTL,Lai2019ExtendingED,Huang2020SemisupervisedNE} have used transfer learning to extend traditional event extraction models to new types without explicitly deriving schemas of new event types.
Nevertheless, these methods still require many annotations for a set of seen types.

In this work, we study the problem of \emph{event type induction} which aims to discover a set of salient event types based on a given corpus.
We observe that about 90\% of event types can be frequently triggered by predicate verbs (c.f. Table~\ref{table:verb_trigger_stats}) and thus propose to take a \emph{verb-centric view} toward inducing event types.
We use the five sentences (S1-S5) in Figure~\ref{fig:intro-example} to motivate our design of event type representation.
First, we observe that verb lemma itself might be ambiguous. For example, the two mentions of lemma \mquote{arrest} in S2 and S3 have different senses and indicate different event types.
Second, even for predicates with the same sense, their different associated object heads\footnote{\small Intuitively, the object head is the most essential word in the object such as \mquote{people} in object \mquote{hundreds of people}.} could lead them to express different event types.
Taking S4 and S5 as examples, two \mquote{stop} mentions have the same sense but belong to different types because of their corresponding object heads. 
Finally, we can see that people have multiple ways to communicate the same event type due to the language variability.

 \begin{table}[!t]
 	\centering
 	\scalebox{0.68}{
         \begin{tabular}{lccc}
         	\toprule
         	Datasets & ACE & ERE & RAMS \\
 		\midrule
		\# of All Event Types & 33 & 38 & 138 \\
		\# of Verb Triggered Event Types & 33 & 38 & 133  \\
		\# of Verb Frequently Triggered Event Types & 28 & 36 & 124  \\
 		\bottomrule
          \end{tabular}
  	}
    \vspace{-0.1cm}  	
	\caption{Statistics of verb triggered event types in three popular event extraction datasets. Event types triggered by verbs more than 5 times are considered as ``Verb Frequently Triggered Event Types''.}
 	\label{table:verb_trigger_stats}
 	\vspace{-0.2cm}
 \end{table}
 
From the above observations, we propose to represent an event type as a cluster of \pair{predicate sense, object head} pairs (P-O pairs for short)\footnote{\small Subjects are intentionally left here because \cite{Allerton1979EssentialsOG} finds objects play a more important role in determining predicate semantics. Also, many P-O pairs indicate the same event type but share different subjects (\eg, \mquote{police capture X} and \mquote{terrorists capture X} are considered as two different events but belong to the same event type \texttt{Capture Person}. Adding subjects may help divide current event types into more fine-grained types and we leave this for future work.}.
We present a new event type induction framework \textbf{\textsc{ETypeClus}} to automatically discover event types, customized for a specific input corpus. 
\ours requires no human-labeled data other than an existing general-domain verb sense dictionary such as VerbNet~\cite{Schuler2005VerbnetAB} and OntoNotes Sense Groupings~\cite{Hovy2006OntoNotesT9}.
\ours contains four major steps.
First, it extracts \pair{predicate, object head} pairs from the input corpus based on sentence dependency tree structures.
As some extracted pairs could be too general (\eg, \pair{say, it}) or too specific (\eg, \pair{document, microcephaly}), the second step of \ours will identify salient predicates and object heads in the corpus.
After that, we disambiguate the sense of each predicate verb by comparing its usage with those example sentences in a given verb sense dictionary.
Finally, \ours clusters the remaining salient P-O pairs into event types using a latent space generative model.
This model jointly embeds P-O pairs into a latent spherical space and performs clustering within this space. 
By doing so, we can guide the latent space learning with the clustering objective and enable the clustering process to benefit from the well-separated structure of the latent space. 

We show our \ours framework can save annotation cost and output corpus-specific event types on three datasets. 
The first two are benchmark datasets ACE 2005 and ERE (Entity Relation Event)~\cite{Song2015FromLT}. 
\ours can successfully recover predefined types and identify new event types such as \texttt{Build} in ACE and \texttt{Bombing} in ERE. 
Furthermore, to test the performance of \ours in new domains, we collect a corpus about the disease outbreak scenario.
Results show that \ours can identify many interesting fine-grained event types (\eg, \texttt{Vaccinate}, \texttt{Test}) that align well with human annotations. 

\smallskip
\noindent \textbf{Contributions.} The major contributions of this paper are summarized as follows: 
(1) A new event type representation is created as a cluster of \pair{predicate sense, object head} tuples;
(2) a novel event type induction framework \ours is proposed that automatically disambiguates predicate senses and learns a latent space with desired event cluster structures; and 
(3) extensive experiments on three datasets verify the effectiveness of \ours in terms of both automatic and human evaluations.


\section{Problem Formulation}\label{sec:problem}

\begin{figure*}[!t]
	\centering
	\centerline{\includegraphics[width=0.95\textwidth]{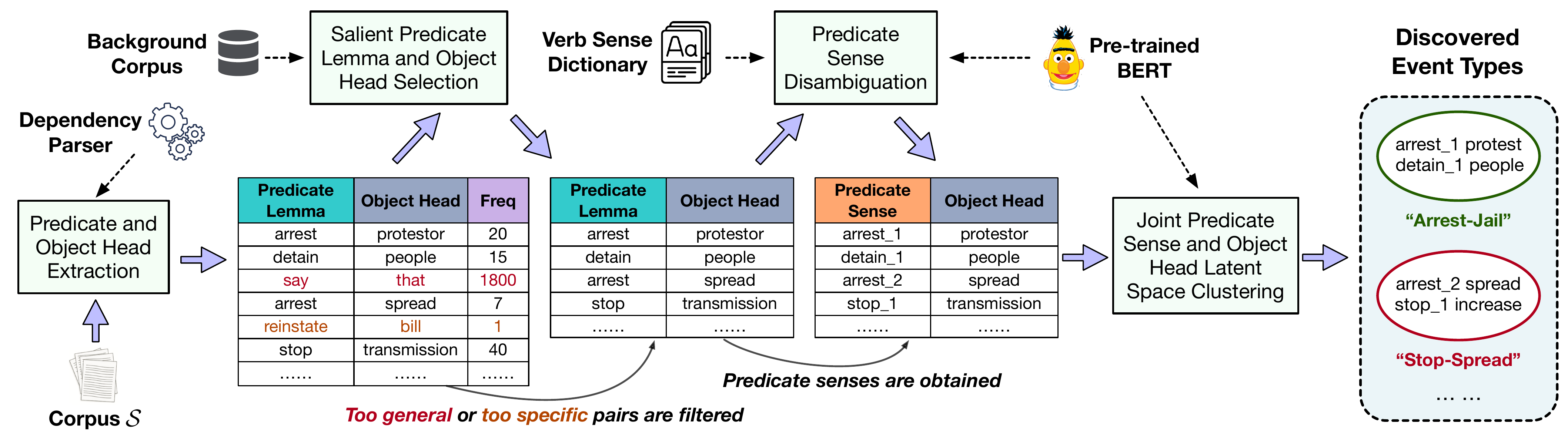}}
	\vspace{-0.2cm}
	\caption{Our \ours framework overview.}
	\label{fig:framework}
	\vspace{-0.3cm}
\end{figure*}

In this section, we first introduce some important concepts and then present our task definition.
A \textbf{corpus} $\mathcal{S} = \{S_1, \dots, S_{N} \}$ is a set of sentences where each sentence $S_i \in \mathcal{S}$ is a word sequence $[w_{i,1}, \dots, w_{i,n}]$.
A \textbf{predicate} is a \emph{verb mention in a sentence} and can optionally have an associated \textbf{object} in the same sentence. 
We follow previous studies~\cite{corbett1993heads,OGorman2016RicherED} and refer to the most important word in the object as the \textbf{object head}.
For example, one predicate from the first sentence in Figure~\ref{fig:intro-example} is \mquote{detain} and its corresponding object is \mquote{hundreds of people} with the word \mquote{people} being the object head.

\begin{figure}[!t]
	\centering
	\centerline{\includegraphics[width=0.46\textwidth]{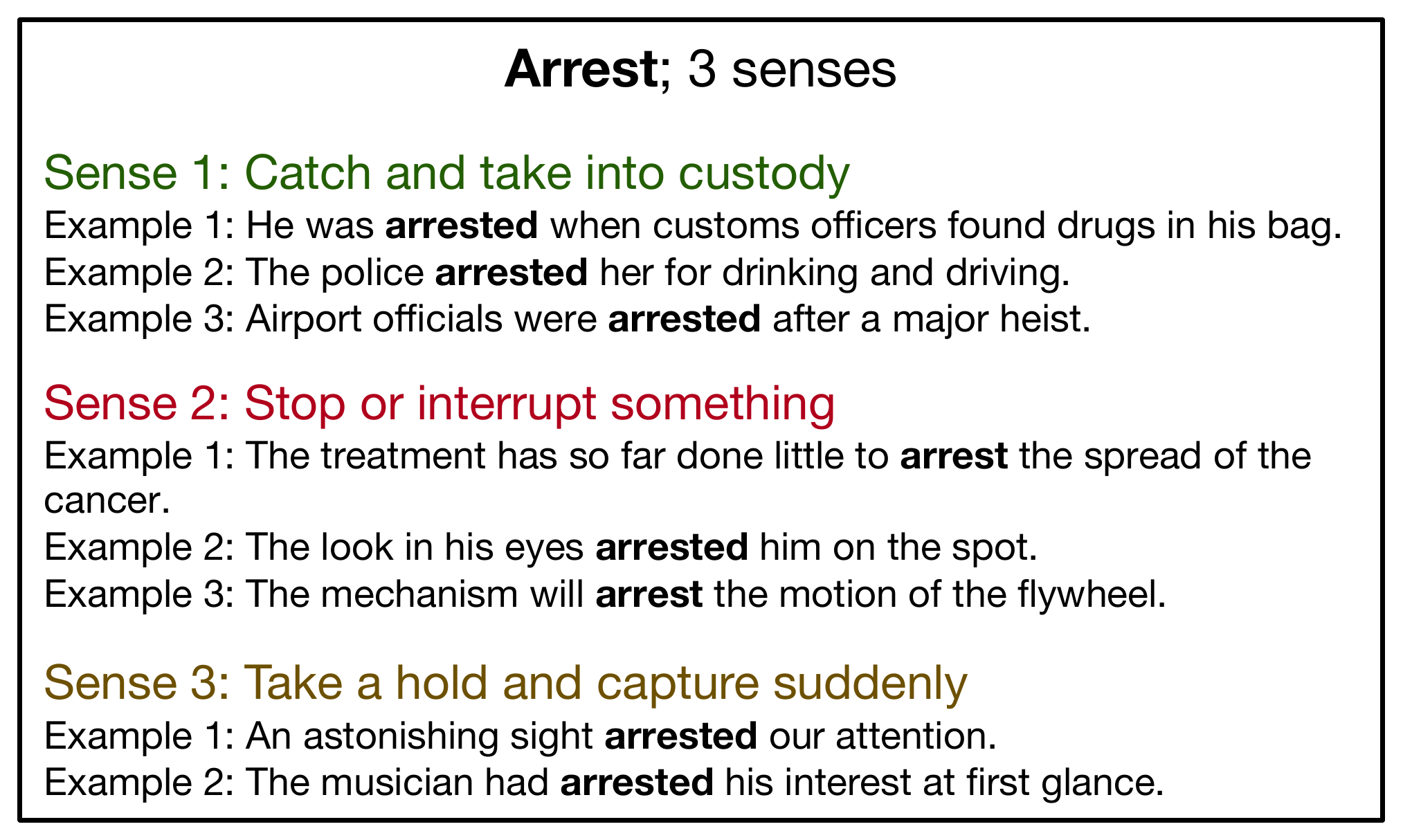}}
	\vspace{-0.2cm}
	\caption{One example in verb sense dictionary $\mathcal{V}$.}
	\label{fig:verb-dict}
	\vspace{-0.3cm}
\end{figure}

As predicates with the same lemma may have different senses, we disambiguate each predicate verb based on a \textbf{verb sense dictionary} $\mathcal{V}$ wherein each verb lemma has a list of candidate senses with example usage sentences.
One illustrative example of our verb sense dictionary is shown in Figure~\ref{fig:verb-dict}.
We refer to the sense of predicate verb lemma as the \textbf{predicate sense}. 

\smallskip
\noindent \textbf{Task Definition.}
Given a corpus $\mathcal{S}$ and a verb sense dictionary $\mathcal{V}$, our task of \emph{event type induction} is to identify a set of $K$ event types where each type $T_{j}$ is represented by a cluster of \pair{predicate sense, object head} pairs.

\section{The \ours Framework}\label{sec:method}

The \ours framework (outlined in Figure~\ref{fig:framework}) induces event types in four major steps: 
(1) predicate and object head extraction,
(2) salient predicate lemma and object head selection,
(3) predicate sense disambiguation, and
(4) latent space joint predicate sense and object head clustering.

\subsection{Predicate and Object Head Extraction}\label{subsec:poh_extraction}

We propose a lightweight method to extract predicates and object heads in sentences without relying on manually-labeled training data.
Specifically, given a sentence $S_i$, we first use a dependency parser\footnote{\small We use the Spacy \texttt{en\_core\_web\_lg} model.} to obtain its dependency parse tree and select all non-auxiliary verb tokens\footnote{\small A token with part-of-speech tag \texttt{VERB} and dependency label not equal to \texttt{aux} and \texttt{auxpass}.} as our candidate predicates.
Then, for each candidate predicate, we check its dependent words and if any of them has a dependency label \texttt{auxpass}, we believe this predicate verb is in passive voice and find its object heads within its syntactic children that occur before it and have a dependency label in SUBJECT label set\footnote{\small \{\texttt{nsubj(pass)}, \texttt{csubj(pass)}, \texttt{agent}, \texttt{expl}\}}.
Otherwise, we consider this predicate is in active voice and identify its object heads within its dependents that occur after it and have a dependency label in OBJECT label set\footnote{\footnotesize \{\texttt{dobj}, \texttt{dative}, \texttt{attr}, \texttt{oprd}\}}.
Finally, we aggregate all \pair{predicate, object head} pairs along with their frequencies in the corpus.


\subsection{Salient Predicate Lemma and Object Head Selection}\label{subsec:ploh_selection}

The above extracted \pair{predicate, object head} pairs have different qualities.
Some are too general and contain little information, while others are too specific and hard to generalize. 
Thus, this step of \ours tries to select those salient predicate lemmas and object heads from our input corpus.

We compute the salience of a word (either a predicate lemma or an object head) based on two criteria. 
First, it should appear frequently in our corpus.
Second, it should not be too frequent in a large general-domain background corpus\footnote{\small We use the English Wikipedia 20171201 dump as our background corpus.}. 
Computationally, we follow the TF-IDF idea and define the word salience as follows:
\begin{equation}
\small
Salience(w) = \left(1+\log(freq(w))^{2}\right) \log(\frac{N\_bs}{bsf(w)}),
\end{equation}
where $freq(w)$ is the frequency of word $w$, $N\_bs$ is the number of \underline{b}ackground \underline{s}entences, and $bsf(w)$ is the \underline{b}ackground \underline{s}entence \underline{f}requency of word $w$. 
Finally, we select those terms with salience scores ranked in top 80\% as our salient predicate lemmas and object heads.
Table~\ref{table:example_saliency} lists the top 5 most salient predicate lemmas and object heads in three datasets.
The first two datasets contain news articles about wars and thus terms like \mquote{kill} and \mquote{weapon} are ranked top.
The third dataset includes articles about disease outbreaks and thus most salient terms include \mquote{infect}, \mquote{virus}, and \mquote{outbreak}.

\subsection{Predicate Sense Disambiguation}\label{subsec:psd}

As verbs typically exhibit large sense ambiguities, we disambiguate each predicate's sense in the sentence.
\citet{Huang2016LiberalEE} achieves this goal by utilizing a supervised word sense disambiguation tool~\cite{Zhong2010ItMS} to link each predicate to a WordNet sense~\cite{Miller1995WordNetAL} and then mapping that sense back to an OntoNotes sense grouping~\cite{Hovy2006OntoNotesT9}. 
In this work, we propose to remove such extra complexity and present a lightweight sense disambiguation method that requires only a verb sense dictionary.

The key idea of our method is to compare the usage of a predicate with each verb sense's example sentences in the dictionary.
Given a predicate verb $v$ in sentence $S_i$, we compute two types of features to capture both its \emph{content} and \emph{context} information. 
The first one, denoted as $\mathbf{v}^{emb}$, is obtained by feeding the sentence $S_i$ into the BERT-Large model~\cite{Devlin2019BERTPO} and retrieving the predicate's corresponding contextualized embedding.
The second feature $\mathsf{v}^{mwp}$ is a rank list of 10 alternative words that can be used to replace $v$ in sentence $S_i$.
Specifically, we replace the original word $v$ in $S_i$ with a special \texttt{[MASK]} token and feed the masked sentence $S_{i}^{mask}$ into BERT-Large for masked word prediction.
From the prediction results, we select the top 10 most likely words and sort them into $\mathsf{v}^{mwp}$.

 \begin{table}[!t]
 	\centering
 	\scalebox{0.75}{
         \begin{tabular}{cc|cc|cc}
         	\toprule
         	 \multicolumn{2}{c}{\textbf{ACE}} & \multicolumn{2}{c}{\textbf{ERE}} & \multicolumn{2}{c}{\textbf{Pandemic}} \\
                  	PredL & ObjH & PredL & ObjH & PredL & ObjH  \\
 		\midrule
		kill  			& weapon		& pay 	& money 	& infect 		& virus		\\
		pay			& iraqis 		& kill 	& people 	& suspect		& outbreak	\\
		guess 		& nations 		& rape 	& kid 	& sicken		& vaccine		\\   
		convict 		& states		& send 	& weapon & test		& case		\\   
		fire  			& marines		& attack 	& cadre 	& circulate	 	& infection		\\
 		\bottomrule
          \end{tabular}
  	}
	\caption{Top 5 salient predicate lemmas (PredL) and object heads (ObjH) in three datasets.}
 	\label{table:example_saliency}
 	\vspace{-0.1cm}
 \end{table}
 
After obtaining the predicate representation, we compute the representations of its candidate senses in the dictionary.
Suppose the lemma of this predicate $v$ has $N_{v}$ candidate senses in the dictionary and each sense $E_{j}$, $j \in [1, \dots, N_{v}]$ has $N_j$ example sentences $\{S_{j,k}\}_{k=1}^{N_j}$ in the dictionary.
Then,  within each example sentence $S_{j,k}$, we locate where the predicate lemma $v$ occurs and compute its corresponding feature $\mathbf{v}^{emb}_{j,k}$ and $\mathsf{v}^{mwp}_{j,k}$ similarly as discussed before.
After that, we obtain two types of features for each sense $E_j$ as follows:
\begin{equation}
\small
\mathbf{E}^{emb}_{j} = \frac{1}{N_j} \sum_{k=1}^{N_j} \mathbf{v}^{emb}_{j,k}, ~~
\mathsf{E}^{mwp}_{j} = RA(\{\mathsf{v}^{mwp}_{j,k}\}|_{k=1}^{N_j}),
\end{equation}
where $RA(\cdot)$ stands for the rank aggregation operation based on mean reciprocal rank.
This method is widely used in previous literature~\cite{Shen2017SetExpanCS,Shen2020SynSetExpanAI,Zhang2020EmpowerES,Huang2020GuidingCS} for fusing ranked lists.
Finally, we choose the sense that is most similar to the predicate $v$ as follows:
\begin{equation}
\small
j^{*} = \argmax_{j\in [1, \dots, N_{v}]}~\text{cos}(\mathbf{v}^{emb}, \mathbf{E}^{emb}_{j}) \cdot \text{rbo}(\mathsf{v}^{mwp}, \mathsf{E}^{mwp}_{j}),
\end{equation}
where $\text{cos}(\mathbf{x}, \mathbf{y})$ is the cosine similarity between two vectors $\mathbf{x}$ and $\mathbf{y}$, and $\text{rbo}(\mathsf{a}, \mathsf{b})$ is the rank-biased overlap similarity~\cite{Webber2010ASM} between two ranked lists.

We evaluate our method on the verb subset of standard word sense disambiguation benchmarks~\cite{Navigli2017WordSD}.
Our method achieves 55.7\% F1 score.
In comparison, the supervised IMS method in~\cite{Huang2016LiberalEE} gets a 56.9\% F1 score.
Thus, we think our method is comparable to supervised IMS but being more lightweight and requires no training data.

\subsection{Latent Space Joint Predicate Sense and Object Head Clustering}\label{subsec:joint_clus}

After obtaining salient \pair{predicate sense, object head} pairs (P-O pairs for short), we aim to cluster them into event types.
Below, we first discuss how to obtain the initial features for predicate senses and object heads (Section~\ref{subsubsec:initial_feature}). 
As those predicate senses and object heads are living in two separate spaces, we aim to fuse them into one joint feature space wherein the event cluster structures are better preserved.
Inspired by \cite{meng2022topic}, we achieve this goal by proposing a latent space generative method that jointly embeds P-O pairs into a unified spherical space and performs clustering in this space. 
Finally, we discuss how to train this generative model in Section~\ref{subsubsec:model_train}.

\subsubsection{Initial Feature Acquisition}\label{subsubsec:initial_feature}

We obtain two types of features for each term $w$ (either a predicate sense $w_p$ or an object head $w_o$) by first locating its mentions in the corpus and then aggregating mention-level representations into term-level features.
Suppose term $w$ appears $M_w$ times, for each of its mentions $m_{w,l}, l \in [1, \dots, M_w]$,  we extract this mention's content feature $\mathbf{m}^{emb}_{w,l}$ and context feature $\mathsf{m}^{mwp}_{w,l}$, following the same process discussed in Section~\ref{subsec:psd}.
Then, we average all mentions' content features into this term's content feature $\mathbf{m}^{emb}_{w} = \frac{1}{M_w} \sum_{l=1}^{M_w} \mathbf{m}^{emb}_{w,l}$.

The aggregation of mention context features is more difficult as each $\mathsf{m}^{mwp}_{w,l}$ is not a numerical vector but instead a set of words predicted by BERT to replace $m_{w,l}$.
In this work, we propose the following aggregation scheme.
For each term $w$, we first construct a pseudo document $D_{w}$ using the bag union operation\footnote{\small Namely, $D_{w}$ contains a word $T$ times if this word appears in $T$ different $\mathsf{m}^{mwp}_{w,l}$, $l \in [1, \dots, M_w]$.}.
Then, we obtain the vector representations of pseudo documents based on TF-IDF transformation and apply Principal Component Analysis (PCA) to reduce the dimensionality of document vectors.
A similar idea is discussed before in~\cite{Amrami2018WordSI}.
The resulting vector will be considered as the term's context feature vector $\mathbf{m}^{mwp}_{w}$.
Finally, we concatenate $\mathbf{m}^{emb}_{w}$ with $\mathbf{m}^{mwp}_{w}$ to obtain the initial feature vector of predicate senses (denoted as $\mathbf{h}_{p}$) and object heads (denoted as $\mathbf{h}_{o}$).

\subsubsection{Latent Space Generative Model}\label{subsubsec:gen_model}
To cluster P-O pairs into $K$ event types based on two separate feature spaces ($\mathbf{H}_p$ for predicate sense and $\mathbf{H}_o$ for object head), one straightforward approach is to represent each P-O pair $x=(p, o)$ as $\mathbf{x} = [\mathbf{h}_p, \mathbf{h}_o]$ and directly applying clustering algorithms to all pairs.
However, this approach cannot guarantee the concatenated space $\mathbf{H} = [\mathbf{H}_p, \mathbf{H}_o]$ will be naturally suited for clustering.
Therefore, we propose to jointly embed and cluster P-O pairs in latent space $\mathbf{Z}$.
By doing so, we can unify two feature spaces $\mathbf{H}_p$ and $\mathbf{H}_o$.
More importantly, the latent space learning is guided by the clustering objective, and the clustering process can benefit from the well-separated structure of the latent space, which achieves a mutually-enhanced effect.

We design the latent space to have a spherical topology because cosine similarity more naturally captures word/event semantic similarities than Euclidean/L2 distance.
Previous studies~\cite{Meng2019SphericalTE,Meng2020HierarchicalTM} also show that learning spherical embeddings directly is better than first learning Euclidean embeddings and normalizing them later.
Thus, we assume there is a spherical latent space $\mathbf{Z}$ with $K$ clusters\footnote{\small $K$ is a hyper-parameter. We can either set $K$ to the true number of event types (if it is known) or directly set $K$ based on application-specific knowledge or adopt statistical methods to estimate $K$. In practice, we can set it to a relatively high number and the resulting event types are still useful.}.
Each cluster in this space corresponds to one event type and is associated with a von Mises-Fisher (vMF) distribution~\cite{Banerjee2005ClusteringOT} from which event type representative P-O pairs are generated.
The vMF distribution of an event type $c$ is parameterized by a mean vector $\mathbf{c}$ and a concentration parameter $\kappa$.
A unit-norm vector $\mathbf{z}$ is generated from $\text{vMF}_{d}(\mathbf{c}, \kappa)$ with the probability as follows:
\begin{equation}
\small
p(\mathbf{z}|\mathbf{c}, \kappa) = n_{d}(\kappa) \exp(\kappa \cdot \text{cos}(\mathbf{z}, \mathbf{c})),
\end{equation}
where $d$ is the dimensionality of latent space $\mathbf{Z}$ and $n_{d}(\kappa)$ is a normalization constant. 

\begin{figure}[!t]
	\centering
	\centerline{\includegraphics[width=0.47\textwidth]{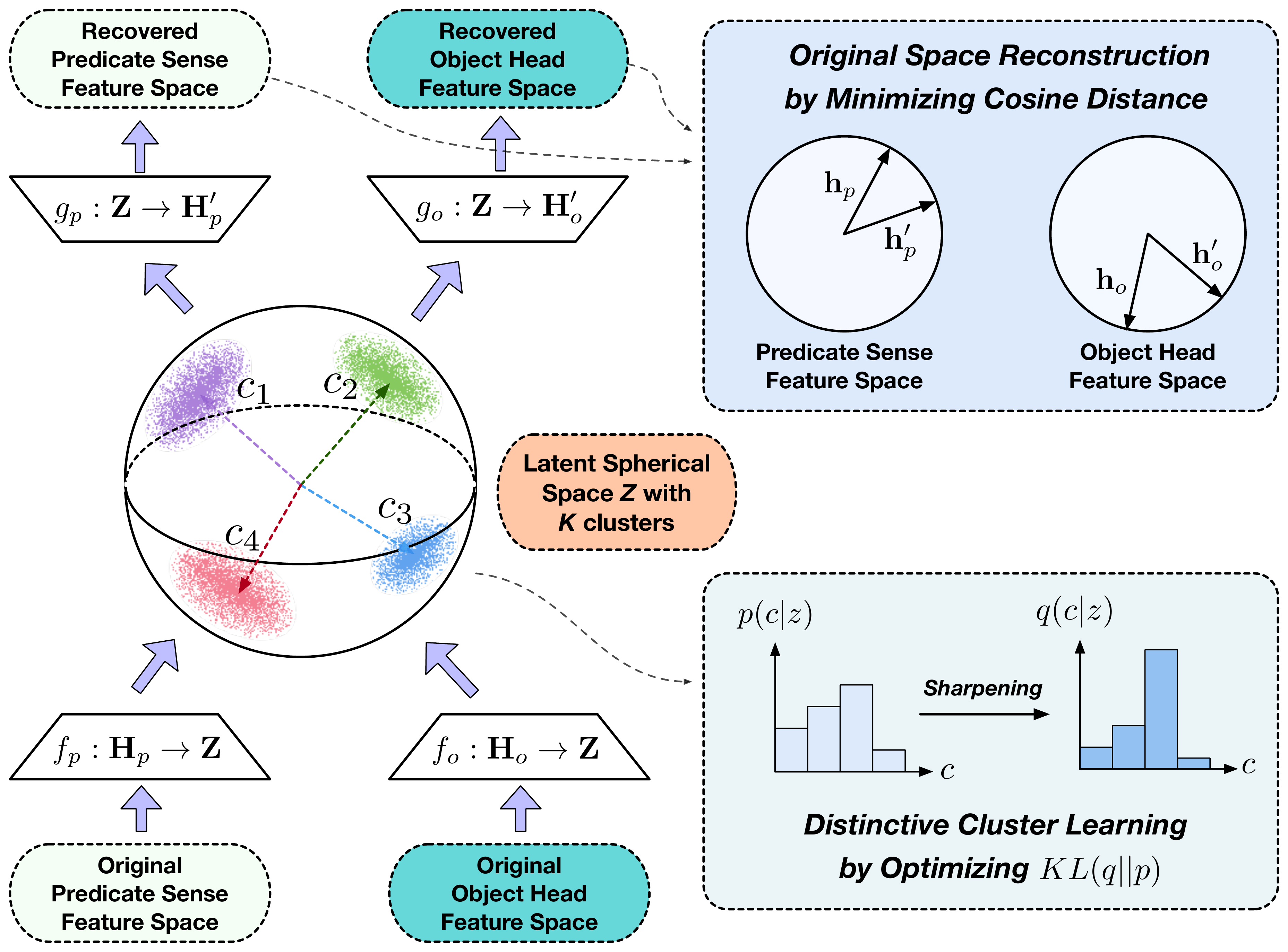}}
	\vspace{-0.2cm}
	\caption{Overview of joint predicate sense and object head latent spherical space clustering. Detailed descriptions in Section~\ref{subsec:joint_clus}.}
	\label{fig:autoencoder}
	\vspace{-0.3cm}
\end{figure}

Each P-O pair $(p_i, o_i)$ with the initial feature $[\mathbf{h}_{p_i}, \mathbf{h}_{o_i}] \in \mathbf{H}_p \times \mathbf{H}_o$ is assumed to be generated as follows:
(1) An event type $c_{k}$ is sampled from a uniform distribution over $K$ types;
(2) a latent embedding $\mathbf{z}_i$ is generated from the vMF distribution associated with $c_{k}$; and
(3) a function $g_p$ ($g_o$) maps the latent embedding $\mathbf{z}_i$ to the original embedding $\mathbf{h}_{p_i}$ ($\mathbf{h}_{o_i}$) corresponding to the predicate sense $p_i$ (object head $o_i$).
Namely, we have:
\begin{equation}
{\small
\begin{split}
c_{k} \sim \text{Uniform}(K)&, \quad \mathbf{z}_i \sim \text{vMF}_{d}(\mathbf{c}_k, \kappa), \\
\mathbf{h}_{p_i} = g_{p}(\mathbf{z}_i)&,\quad \mathbf{h}_{o_i} = g_{o}(\mathbf{z}_i).
\end{split}
}
\end{equation}
We parameterize $g_{p}$ and $g_o$ as two deep neural networks and jointly learn the mapping function $f_p: \mathbf{H}_p \rightarrow \mathbf{Z}$ as well as $f_o: \mathbf{H}_o \rightarrow \mathbf{Z}$ from the original space to the latent space. 
Such a setup closely follows the autoencoder architecture~\cite{Hinton1993AutoencodersMD} which is shown to be effective for preserving input information.

\subsubsection{Model Training}\label{subsubsec:model_train}

We learn our generative model by jointly optimizing two objectives. 
The first one is a \emph{reconstruction objective} defined as follows:
\begin{equation}\label{eq:reconstruction_loss}
\small
\mathcal{O}_{\text{rec}} = \sum_{i=1}^{N} \Big( \text{cos}(\mathbf{h}_{p_i}, g_p(f_p(\mathbf{h}_{p_i})))  + \text{cos}(\mathbf{h}_{o_i}, g_o(f_o(\mathbf{h}_{o_i}))  ) \Big)
\end{equation}
This objective encourages our model to preserve input space semantics and generate the original data faithfully. 

The second \emph{clustering-promoting objective} enforces our model to learn a latent space with $K$ well-separated cluster structures.
Specifically, we use an expectation-maximization (EM) algorithm to sharpen the posterior event type distribution of each input P-O pair.
In the expectation step, we first compute the posterior distribution based on current model parameters as follows:
\begin{equation}\label{eq:e_step}
{\small
\begin{split}
p(c_k|\mathbf{z}_i) & = \frac{p(\mathbf{z}_i | c_k)p(c_k) }{\sum_{k'=1}^{K} p(\mathbf{z}_i | c_{k'})p(c_{k'})} \\
				 & = \frac{\exp(\kappa \cdot \text{cos}(\mathbf{z}_i, \mathbf{c}_k)) }{\sum_{k'=1}^{K} \exp(\kappa \cdot \text{cos}(\mathbf{z}_i, \mathbf{c}_{k'})) }.
\end{split}
}
\end{equation}

\begin{table*}[t]
\small
    \centering
    \scalebox{.765}{
    \begin{tabular}{m{2.0cm} m{3.5cm}m{13.4cm}}
        \toprule
         \textbf{Event Type} & \textbf{Top Ranked P-O Pairs} & \quad \quad    \textbf{Example Sentences in Corpus}\\
         
        \midrule
        \vspace{0.3cm}
         \vbox{\nop{\hbox{\strut Justice:}}\hbox{\strut Arrest-Jail}\nop{\hbox{\strut (ACE)}}}  & \begin{itemize}[label={},leftmargin=-1px]
        \setlength\itemsep{-.5em}
                \vspace{-.3cm}
            \item \pair{arrest\_0, protester}
            \item \pair{arrest\_0, militant}
            \item \pair{arrest\_0, suspect}
            \vspace{-.3cm}
        \end{itemize} &  
        \vspace{-.3cm}
        \begin{itemize}
        \setlength\itemsep{-.5em}
            \item For the most part the marches went off peacefully, but in New York a small group of \emph{\underline{protesters}} were \textbf{arrested} after they refused to go home at the end of their rally, police sources said.
            \item On Tuesday, Saudi security officials said three suspected al-Qaida \emph{\underline{militants}} were \textbf{arrested} in Jiddah, Saudi Arabia.
            \vspace{-.3cm}
        \end{itemize}
\\
        \midrule
        \vspace{0.3cm}
         \vbox{\hbox{\strut Build$^\nabla$}\nop{\hbox{\strut (ACE)}}}  & \begin{itemize}[label={},leftmargin=-1px]
        \setlength\itemsep{-.5em}
                \vspace{-.3cm}
            \item \pair{build\_0, facility}
            \item \pair{build\_0, center}
            \item \pair{build\_0, housing}
            \vspace{-.3cm}
        \end{itemize} &  
        \vspace{-.3cm}
        \begin{itemize}
        \setlength\itemsep{-.5em}
            \item Plans were underway to \textbf{build} destruction \emph{\underline{facilities}} at all other locations but now the Bush junta has removed from its proposed defense budget for fiscal year 2006 all but the minimum funding.
            \item Virginia is apparently going to be \textbf{build} a data \emph{\underline{center}} in Richmond, a back-up data center, and a help desk/call center as a follow-on to the creation of VITA, the Virginia Information Technology Agency.
            \vspace{-.3cm}
        \end{itemize}
\\
        \midrule
        \midrule
        \vspace{0.3cm}
         \vbox{\nop{\hbox{\strut Transaction:}}\hbox{\strut Transfer-Money}\nop{\hbox{\strut (ERE)}}}  & \begin{itemize}[label={},leftmargin=-1px]
        \setlength\itemsep{-.5em}
                \vspace{-.3cm}
            \item \pair{fund\_0, activity}
            \item \pair{fund\_0, operation}
            \item \pair{fund\_0, people}
            \vspace{-.3cm}
        \end{itemize} &  
        \vspace{-.3cm}
        \begin{itemize}
        \setlength\itemsep{-.5em}
            \item The grants will \textbf{fund} advisory \emph{\underline{activities}}, including local capacity building, infrastructure development and product development.
            \item The White House had hoped to hold off asking for more money to \textbf{fund} military \emph{\underline{operations}} in Iraq and Afghanistan until after the election, but with costs rising faster than expected, it sent a request for an early installment of \$25 billion to Congress this week.
            \vspace{-.3cm}
        \end{itemize}
\\
        \midrule
        \vspace{0.3cm}
         \vbox{\hbox{\strut Bombing$^\nabla$}\nop{\hbox{\strut (ERE)}}}  & \begin{itemize}[label={},leftmargin=-1px]
        \setlength\itemsep{-.5em}
                \vspace{-.3cm}
            \item \pair{bomb\_0, factory}
            \item \pair{bomb\_0, checkpoint}
            \item \pair{bomb\_0, base}
            \vspace{-.3cm}
        \end{itemize} &  
        \vspace{-.3cm}
        \begin{itemize}
        \setlength\itemsep{-.5em}
            \item He \textbf{bombed} the Aspirin \emph{\underline{factory}} in 1998 (which turned out to have nothing to do with Bin Laden) the week he revealed he had been lying to us for eight months about Lewinsky.
            \item Prosecutors then also pointed to the men’s suicide bomber training in 2011 in Somalia and association with Beledi, who prosecutors said \textbf{bombed} a government \emph{\underline{checkpoint}} in Mogadishu that year.
            \vspace{-.3cm}
        \end{itemize}
\\
        \bottomrule
        \end{tabular}
    }
    \vspace*{-0.2cm}
    \caption{Example outputs of \ours discovered event types with their associated sentences in ACE and ERE datasets. The first two types come from ACE and the remaining two are from ERE. The event types with superscript ``$^\nabla$'' originally do not exist in human-labeled schemas and are discovered by \ours framework. \textbf{Predicates} are in bold and \underline{\emph{object heads}} are underlined and in italics.}
    \label{table:benchmark_cases}
    \vspace*{-0.3cm}
\end{table*}

We then compute a new estimate of each P-O pair's cluster assignment $q(c_k|\mathbf{z}_i)$ and use it to update the model in the maximization step.
Instead of making hard cluster assignments like K-means which directly assigns each $\mathbf{z}_i$ to its closest cluster, we compute a soft assignment $q(c_k|\mathbf{z}_i)$ as follows:
\begin{equation}\label{eq:q_prob}
\small
q(c_k|\mathbf{z}_i) = \frac{p(c_k|\mathbf{z}_i)^2 / s_k}{\sum_{k'=1}^{K} p(c_{k'}|\mathbf{z}_i)^2 / s_{k'}},
\end{equation}
where $s_{k} = \sum_{i=1}^{N} p(c_k|\mathbf{z}_i)$. 
This squaring-then-normalizing formulation has a sharpening effect that skews the distribution towards its most confident cluster assignment, as shown in~\cite{Xie2016UnsupervisedDE,Meng2018WeaklySupervisedNT,Meng2019WeaklySupervisedHT}.
The formulation encourages unambiguous assignment of P-O pairs to event types so that the learned latent space will have gradually well-separated cluster structures.
Finally, in the maximization step, we update the model parameters to maximize the expected log-probability of the current cluster assignments under the new cluster assignment estimates as follows:
\begin{equation}\label{eq:cluster_loss}
\small
\mathcal{O}_{\text{clus}} = \sum_{i=1}^{N}  \sum_{k=1}^{K} q(c_k|\mathbf{z}_i) \log p(c_k|\mathbf{z}_i),
\end{equation}
where $p$ is updated to approximate fixed target $q$.

 \small
 \begin{algorithm}[!t]
  \caption{Latent Space Generative Model Training.}
  \label{algo:train_cluster}
  \KwIn{
  \small A set of P-O pairs $\{x_i\}_{i=1}^{N}$; Initial feature spaces $\mathbf{H}_p$ and $\mathbf{H}_o$; \# of event types $K$.
  }
  \KwOut{\small Event-pair distributions $p(x_i|c_k)$.}
  \small $f_o, f_p, g_o, g_p \leftarrow \text{max}~\mathcal{O}_{\text{rec}}$ in Eq.~(6) ~~//~Pretraining\; 
  \small Initialize $\mathbf{C} = \{\mathbf{c}_{k}\}_{k=1}^{K}$\;
  \small \While{not converaged} {
  	\small //~Update cluster assignment estimation\;
	\small $q(c_k|\mathbf{z}_i) \leftarrow$ Eq.~(8)\; 
	\small //~Update model parameteres\;
	\small $f_o, f_p, g_o, g_p, \mathbf{C} \leftarrow \text{max}~\mathcal{O}_{\text{rec}} + \lambda \mathcal{O}_{\text{clus}}$\;
  }
  \small Return $p(x_i|c_k) = p(\mathbf{z}_i|\mathbf{c}_k)$\;   
 \end{algorithm}
 \normalsize
 
We summarize our training procedure in Algorithm~\ref{algo:train_cluster}.
We first pretrain the model using only the reconstruction objective, which provides a stable initialization of all parameterized mapping functions.
Then, we apply the EM algorithm to iteratively update all mapping functions and event type parameters $\mathbf{C}$ with a joint objective $\mathcal{O}_{\text{rec}} + \lambda \mathcal{O}_{\text{clus}}$ where the hyper-parameter $\lambda$ balances two objectives.
The algorithm is considered converged if less than $\delta=5\%$ of the P-O pairs change cluster assignment between two iterations or a maximum iteration number is reached.
Finally, we output each P-O pair's distribution over $K$ event types.

\section{Evaluation on ACE/ERE Datasets}\label{sec:exp_ace}

\begin{table*}[t]
\centering
\scalebox{0.61}{
\begin{tabular}{c|cccc|cccc}
\toprule
\multirow{2}{*}{Methods}    & \multicolumn{4}{c}{ACE}                  & \multicolumn{4}{c}{ERE}                     \\
\cmidrule{2-5} \cmidrule{6-9}
                                                & ARI (std)    & NMI (std)    & ACC (std)    & BCubed-F1 (std)& ARI (std)    & NMI (std)    & ACC (std)    & BCubed-F1 (std) \\
\midrule
Kmeans                                          & 26.27 (1.60) & 48.02 (1.55) & 41.57 (3.07) & 41.33 (1.75)   & 11.17 (1.83) & 35.10 (2.36) & 31.65 (1.82) & 29.97 (1.79) \\
sp-Kmeans                                       & 26.06 (2.12) & 47.30 (1.65) & 40.41 (2.46) & 39.52 (1.42)   & 13.62 (2.14) & 37.33 (2.25) & 33.28 (3.12) & 30.73 (2.03) \\
AggClus                                    & 24.45 (0.00) & 45.71 (0.00) & 41.00 (0.00) & 40.20 (0.00)   & 6.07 (0.00)  & 29.62 (0.00) & 30.84 (0.00) & 29.90 (0.00) \\
Triframes~\cite{Ustalov2018UnsupervisedSF}      & 19.35 (6.60) & 36.38 (4.91) & ---          & 38.91 (2.36)   & 10.89 (2.51) & 34.94 (2.54) & ---          & 33.53 (4.47) \\
JCSC~\cite{Huang2016LiberalEE}                  & 36.10 (4.96) & 49.50 (2.70) & 46.17 (3.64) & 43.83 (3.17)   & 17.07 (4.40) & 39.50 (3.97) & 33.76 (2.43) & 34.04 (2.23) \\
\ours                                           & \bf 40.78 (3.20) & \bf 57.57 (2.40) & \bf 48.35 (2.55) & \bf 51.58 (2.50)   & \bf 24.09 (1.93) & \bf 49.40 (1.37) & \bf 41.19 (1.87) & \bf 39.78 (1.45) \\   
\bottomrule
\end{tabular}
}
\vspace*{-0.1cm}
\caption{Event mention clustering results. All values are in percentage. We run each method 10 times and report its averaged result for each metric with the standard deviation. Note that ACC is not applicable for Triframes because it assumes the equal number of clusters in ground truth and generated results. 
}
\label{tab:clustering}
\vspace*{-0.3cm}
\end{table*}

We first evaluate \ours on two widely used event extraction datasets: ACE (Automatic Content Extraction) 2005\footnote{\small \url{https://www.ldc.upenn.edu/collaborations/past-projects/ace}} and ERE (Entity Relation Event)~\cite{Song2015FromLT}.
For both datasets, we follow the same preprocessing steps from~\cite{Lin2020AJN,Li2021DocumentLevelEA} and use sentences in the training split as our input corpus. 
The ACE dataset contains 17,172 sentences with 33 event types and the ERE dataset has 14,695 sentences with 38 types.
We test the performance of \ours on event type discovery and event mention clustering.

\begin{table*}[t]
\small
    \centering
    \scalebox{.87}{
    \begin{tabular}{m{1.7cm} m{3.3cm}m{11.5cm}}
        \toprule
         \textbf{Event Type} & \textbf{Top Ranked P-O Pairs} & \quad \quad    \textbf{Example Sentences in Corpus}\\
        \midrule
        \vspace{0.3cm}
         \vbox{\hbox{\strut Spread}\hbox{\strut Virus}}  & \begin{itemize}[label={},leftmargin=-1px]
        \setlength\itemsep{-.5em}
                \vspace{-.3cm}
            \item \pair{spread\_2, virus}
            \item \pair{spread\_2, disease}
            \item \pair{spread\_2, coronavirus}
            \vspace{-.3cm}
        \end{itemize} &  
        \vspace{-.3cm}
        \begin{itemize}
        \setlength\itemsep{-.5em}
            \item What is the best way to keep from \textbf{spreading} the \emph{\underline{virus}} through coughing or sneezing?
            \item Farmers quickly mobilized to fight the misperceptions that pigs could \textbf{spread} the \emph{\underline{disease}}.
            \item In the UK, Asians have been punched in the face, accused of \textbf{spreading} \emph{\underline{coronavirus}}.
            \vspace{-.3cm}
        \end{itemize}
        
\\
	\midrule
        \vbox{\hbox{\strut Prevent}\hbox{\strut Spread}}  & \begin{itemize}[label={},leftmargin=-1px]
        \setlength\itemsep{-.5em}
                \vspace{-.3cm}
            \item \pair{prevent\_1, spread}
            \item \pair{mitigate\_1, spread}
            \item \pair{mitigate\_1, transmission}
            \vspace{-.3cm}
        \end{itemize} &  
        \vspace{-.3cm}
        \begin{itemize}
        \setlength\itemsep{-.5em}
            \item Infection prevention and control measures are critical to \textbf{prevent} the possible \emph{\underline{spread}} of MERS-CoV.
            \item A vaccine can \textbf{mitigate} \emph{\underline{spread}}, but not fully prevent the virus circulating.
            \item Asymptomatic infection could also potentially be directly harnessed to \textbf{mitigate} \emph{\underline{transmission}}.
            \vspace{-.3cm}
        \end{itemize}

\\
	\midrule
        \vbox{\hbox{\strut Vaccinate}\hbox{\strut People}}  & \begin{itemize}[label={},leftmargin=-1px]
        \setlength\itemsep{-.5em}
                \vspace{-.3cm}
            \item \pair{vaccinate\_0, person}
            \item \pair{immunize\_0, people}
            \item \pair{vaccinate\_0, family}
            \vspace{-.3cm}
        \end{itemize} &  
        \vspace{-.3cm}
        \begin{itemize}
        \setlength\itemsep{-.5em}
            \item All \emph{\underline{persons}} in a recommended vaccination target group should be \textbf{vaccinated} with the 2009 H1N1 monovalent vaccine and the seasonal influenza vaccine.
            \item U.K. Will Start \textbf{Immunizing} \emph{\underline{People}} Against COVID-19 On Tuesday, Officials Say.
            \item ``...'' says Henrietta Aviga, a nurse travelling around villages to \textbf{vaccinate} and educate \emph{\underline{families}}.
            \vspace{-.3cm}
        \end{itemize}

         \\
        \bottomrule
        \end{tabular}
    }
    \vspace*{-0.2cm}
    \caption{Example outputs of \ours discovered event types with their associated sentences in the corpus. \textbf{Predicates} are in bold and \underline{\emph{object heads}} are underlined and in italics.
    }
    \label{table:pandemic_cases}
    \vspace*{-0.3cm}
\end{table*}

\begin{table}[t]
\centering
\scalebox{0.76}{
\begin{tabular}{c|cccc}
\toprule
Methods & K-Menas & AggClus & JCSC & \ours \\
\midrule
Accuracy & 86.7 &  64.4 & 54.4 & \bf 91.1 \\
\bottomrule
\end{tabular}
}
\vspace*{-0.2cm}
\caption{Intrusion test results in percentage.}
\label{tab:intrusion}
\vspace*{-0.3cm}
\end{table}    

\subsection{Event Type Discovery}\label{subsec:ace_ere_type_discover}
We apply \ours on each input corpus to discover 100 candidate event clusters and follow~\cite{Huang2016LiberalEE} to manually check whether discovered clusters can reconstruct ground truth event types.
On ACE, we recover 24 out of 33 event types (19 out of 20 most frequent types) and 7 out of 9 missing types have a frequency less than 10.
On ERE, we recover 28 out of 38 event types (18 out of 20 most frequent types). 
We show some example clusters in Table~\ref{table:benchmark_cases} which includes top ranked P-O pairs and their occurring sentences.
We observe that \ours successfully identifies human defined event types (\eg, \texttt{Arrest-Jail} in ACE and \texttt{Transfer-Money} in ERE). 
It can also identify finer-grained types compared with the original ground truth types
(e.g., the 4th row of Table~\ref{table:benchmark_cases} shows one discovered event type \texttt{Bombing} in ERE which is in finer scale than ``Conflict:Attack'', the closest human-annotated type in ERE).
Further, \ours is able to identify new salient event types (e.g., finding new event type \texttt{Build} in ACE).
Finally, \ours not only induces event types but also provides their example sentences, which serve as the \emph{corpus-specific} annotation guidance.

\subsection{Event Mention Clustering}\label{subsec:ace_ere_mention_clus}

We evaluate the effectiveness of our latent space generative model via the event mention clustering task.
We first match each event mention with one extracted P-O pair if possible, and select 15 event types with the most matched results\footnote{\small More details are discussed in Appendix Section D.}. 
Then, for each selected type, we collect its associated mentions and add them into a candidate pool. 
We represent each mention using the feature of its corresponding P-O pair.
Finally, we cluster all mentions in the candidate pool into 15 groups and evaluate whether they align well with the original 15 types.

The event mention clustering quality also serves as a good proxy of the event type quality.
This is because if a method can discover good event types from a corpus, it should also be able to generate good event mention clusters when the ground truth number of clusters is given.

\smallskip
\noindent \textbf{Compared Methods.}
We compare the following methods:
(1) \textbf{Kmeans}: A standard clustering algorithm that works in the Euclidean feature space. We run this algorithm with the ground truth number of clusters.
(2) \textbf{sp-Kmeans}: A variant of Kmeans that clusters event mentions in a spherical space based on the cosine similarity.
(3) \textbf{AggClus}: A hierarchical agglomerative clustering algorithm with Euclidean distance function and Ward linkage. A stop criterion is set to be reaching the target number of clusters.
(4) \textbf{Triframes}~\cite{Ustalov2018UnsupervisedSF}: A graph-based clustering algorithm that constructs a $k$-NN event mention graph and uses a fuzzy graph clustering algorithm \mbox{\sc WATSET} to generate the clusters.
(5) \textbf{JCSC}~\cite{Huang2016LiberalEE}: A joint constrained spectral clustering method that iteratively refines the clustering result with a constraint function to enforce inter-dependent predicates and objects to have coherent clusters. 
(6) \textbf{\textsc{ETypeClus}}: Our proposed latent space joint embedding and clustering algorithm.
For fair comparison, all methods start with the same $[\mathbf{h}_p, \mathbf{h}_o]$ embeddings as described in Section~\ref{subsubsec:gen_model}.
More implementation details and hyper-parameter choices are discussed in Appendix Sections A and B.

\smallskip
\noindent \textbf{Evaluation Metrics.}
We evaluate clustering results with several standard metrics. 
(1) \textbf{ARI}~\cite{Hubert1985ComparingPA} measures the similarity between two cluster assignments based on the number of pairs in the same/different clusters.
(2) \textbf{NMI} denotes the normalized mutual information between two cluster assignments.
(3) \textbf{BCubed-F1}~\cite{Bagga1998EntityCC} estimates the quality of the generated cluster assignment by aggregating the precision and recall of each element.
(4) \textbf{ACC} measures the clustering quality by finding the permutation function from predicted cluster IDs to ground truth IDs that gives the highest accuracy. 
The math formulas of these metrics are in Appendix Section E.
For all four metrics, the higher the values, the better the model performance. 

\smallskip
\noindent \textbf{Experiment Results.}
Table~\ref{tab:clustering} shows \ours outperforms all the baselines on both datasets in terms of all metrics.
The major advantage of \ours is the latent event space: different types of information can be projected into the same space for effective clustering.
We also observe that JCSC is the strongest among all baselines.
We think the reason is that it uses a joint clustering strategy where event types are defined as predicate clusters and the constraint function enables objects to refine predicate clusters.
Thus, a predicate-centric clustering algorithm can outperform all other baselines, which supports our verb-centric view of events.

\section{Evaluation on Pandemic Dataset}\label{sec:exp_pand}

To evaluate the portability of \ours to a new open domain, we collect a new dataset that includes 98,000 sentences about disease outbreak events\footnote{\small The detailed creation process is in Appendix Section F.}.
We run the top-3 performing baselines and \ours to generate 30 candidate event types and evaluate their quality using intrusion test.
Specifically, we inject a negative sample from other clusters into each cluster's top-5 results and ask three annotators to identify the outlier. 
More details on how we construct the intrusions are in Appendix.
The intuition behind this test is that the annotators will be easier to identify the intruders if the clustering results are clean and tuples are semantically coherent.
As shown in Table~\ref{tab:intrusion}, \ours achieves the highest accuracy among all the baseline methods, indicating that it generates semantically coherent types in each cluster.

Table~\ref{table:pandemic_cases} shows some discovered event types of \ours.\footnote{\small More example outputs are in Appendix Section H.}
Interesting examples include tuples with the same predicate sense but object heads with different granularities (\eg, $\langle$spread\_2, virus$\rangle$ and $\langle$spread\_2, coronavirus$\rangle$ for \texttt{Spread-Virus} type),  tuples with same object head but different predicate senses (\eg, $\langle$prevent\_1, spread$\rangle$, and $\langle$mitigate\_1, spread$\rangle$ for \texttt{Prevent-Spread} type),
and event types with predicate verb lemmas that are not directly linkable to OntoNotes Senses grouping (\eg, \mquote{immunize} and \mquote{vaccinate} for \texttt{Vaccinate} type).

\section{Related Work}\label{sec:related_work}

\noindent \textbf{Event Schema Induction.}
Early studies on event schema induction adopt rule-based approaches~\cite{Lehnert1992UniversityOM,Chinchor1993EvaluatingMU} and classification-based methods~\cite{Chieu2003ClosingTG,Bunescu2004CollectiveIE} to induce templates from labeled corpus.
Later, unsupervised methods are proposed to leverage relation patterns~\cite{Sekine2006OnDemandIE,Qiu2008ModelingCI} and coreference chains~\cite{Chambers2011TemplateBasedIE} for event schema induction.
Typical approaches use probabilistic generative models~\cite{Chambers2013EventSI,Cheung2013ProbabilisticFI,Nguyen2015GenerativeES,Li2020ConnectingTD,Li2021DocumentLevelEA} or ad-hoc clustering algorithms~\cite{Huang2016LiberalEE,Sha2016JointLT} to induce predicate and argument clusters.
In particular, \cite{Liu2019OpenDE} takes an entity-centric view toward event schema induction. 
It clusters entities into semantic slots and finds predicates for entity clusters in a post-processing step.
\cite{Yuan2018OpenSchemaEP} studies the event profiling task and includes one module that leverages a Bayesian generative model to cluster \pair{predicate:role:label} triplets into event types.
These methods typically rely on discrete hand-crafted features derived from bag-of-word text representations and impose strong statistics assumptions; whereas our method uses pre-trained language models to reduce the feature generation complexity and relaxes stringent statistics assumptions via latent space clustering.

\smallskip
\noindent \textbf{Weakly-Supervised Event Extraction.}
Some studies on event extraction~\cite{Bronstein2015SeedBasedET,Ferguson2018SemiSupervisedEE,Chan2019RapidCF} propose to leverage annotations for a few seen event types to help extract mentions of new event types specified by just a few keywords.
These methods reduce the annotation efforts but still require all target new types to be given.
Recently, some studies~\cite{Huang2018ZeroShotTL,Lai2019ExtendingED,Huang2020SemisupervisedNE} use transfer learning techniques to extend traditional event extraction models to new types without explicitly deriving schemas of new event types.
Compared to our study, these methods still require many annotations for a set of seen types and their resulting vector-based event type representations are less human interpretable.
Another related work by~\cite{Wang2019OpenEE} uses GAN to extract events from an open domain corpus.
It clusters \pair{entity:location:keyword:date} quadruples related to the same event rather than finds event types.


\section{Conclusions and Future Work}\label{sec:conclusion}
In this paper, we study the event type induction problem that aims to automatically generate salient event types for a given corpus.
We define a novel event type representation and propose \ours that can extract and select salient predicates and object heads, disambiguate predicate senses, and jointly embed and cluster P-O pairs in a latent space.
Experiments on three datasets show that \ours can recover human curated types and identify new salient event types.
In the future, we propose to explore the following directions: 
(1) improve predicate and object extraction quality with tools of higher semantic richness (\eg, a SRL labeler or an AMR parser);
(2) leverage more information from lexical resources to enhance event representation; and 
(3) cluster objects into argument roles for each discovered event type.

\section*{Acknowledgements}
Research was supported in part by US DARPA KAIROS Program No. FA8750-19-2-1004, SocialSim Program No.  W911NF-17-C-0099, and INCAS Program No. HR001121C0165, NSF IIS-19-56151, IIS-17-41317, and IIS 17-04532, and the Molecule Maker Lab Institute: An AI Research Institutes program supported by NSF under Award No. 2019897. 
Any opinions, findings, and conclusions or recommendations expressed herein are those of the authors and do not necessarily represent the views, either expressed or implied, of DARPA or the U.S. Government.
We want to thank Martha Palmer and Ghazaleh Kazeminejad for the help on VerbNet and OntoNotes Sense Groupings.
We also would like to thank Sha Li, Yu Meng, Lifu Huang for insightful discussions and anonymous reviewers for valuable feedback.

\newpage
\section*{Impact Statement}\label{sec:app_ethics}
Both event extraction and event type induction are standard tasks in NLP. We do not see any significant ethical concerns.
The expected usage of our work is to identify interesting event types from user input corpus such as a set of news articles or a collection of scientific papers.

\bibliographystyle{acl_natbib}
\bibliography{custom}


\newpage
\appendix

\section{\textbf{\textsc{ETypeClus}} Implementation Details}

We use the OntoNotes sense grouping\footnote{\small Available to view at \url{http://verbs.colorado.edu/html_groupings/}} as our input verb sense dictionary.
The background Wikipedia corpus is obtained from~\cite{Shen2020SynSetExpanAI}.
We implement our latent space clustering model using PyTorch 1.7.0 with the Huggingface Library~\cite{Wolf2019HuggingFacesTS}.
To obtain each predicate sense's context information $\mathbf{m}_{w}^{mwp}$ (c.f. Section 3.4.1 in main text), we leverage PCA to reduce the original pseudo-document multi-hot representations into 500-dimension vectors.
The hyper-parameters of our latent space generative model are set as follows: the latent space dimension $d=100$, the DNN hidden dimensions are 500-500-1000 for encoder $f_p$/$f_o$ and 1000-500-500 for decoder $g_p$/$g_o$;
the shared concentration parameter of event type clusters $\kappa=10$, the weight for clustering-promoting object $\lambda=0.02$, the convergence threshold $\delta=0.05$, and the maximum iteration number is 100. 
We learn the generative model using Adam optimizer with learning rate 0.001 and batch size 64.

\section{Baseline Implementation Details}

We implement Kmeans and AggClus based on the Scikit-learn codebase~\cite{scikit-learn}. 
We use $L_2$ distance for both methods. 
For Kmeans, we use k-means++ strategy for model initialization, and each time the result with the best inertia is used within 10 initializations. 
We use ward linkage for AggClus and set the stop criterion to be reaching the target number of clusters. 
For spherical Kmeans, we use an open source implementation\footnote{\small \url{https://github.com/jasonlaska/spherecluster}}.
Similar to Kmeans, we use k-means++ to initialize the model and select the best results among 10 initializations.
For Triframes~\cite{Ustalov2018UnsupervisedSF}, we use its authors' original implementation\footnote{\small \url{https://github.com/uhh-lt/triframes}}, and tune the parameter $k$ in the $k$-NN graph construction step for different tasks and datasets to get a reasonable number of clusters. 
Specifically, we use $k=30$ for the event mention clustering task, which gives us the overall best evaluation results on both ACE and ERE. 
On the Pandemic corpus, we take $k=100$, which generates 35 clusters that contain at least 40 tuples. 
For JCSC, we implement the clustering algorithm based on Algorithm~1 in \cite{Huang2016LiberalEE}. 
The spectral clustering used in JCSC is based on Scikit-learn's implementation, and the label assigning strategy is K-means with 30 random initializations each time.

\section{Running Environment}

We run all experiments on a single cluster with 80 CPU cores and a Quadro RTX 8000 GPU.
The BERT model is moved to the GPU for initial predicate sense and object head feature extraction and it consumes about 11GB GPU memory.
We also train our latent space generative model on GPU and it consumes about 14GB GPU memory.
In principles, \ours should be runnable on CPU.

\section{Event Mention Clustering Dataset}
We create the evaluation dataset for event mention clustering as follows.
First, we select event mentions whose trigger is a single token verb. 
Then, for each selected event mention, we construct a P-O pair by choosing its non-pronoun argument that has some overlap with the object of our extracted \pair{predicate, object} with the same verb trigger. 
After that, we select the top-15 event types with the most matched results for both datasets to avoid types with too few mentions, and their corresponding event mentions are used as ground truth clusters.

\section{Evaluation Metrics for Event Mention Clustering}

We denote the ground truth clusters as $C^*$, the predicted clusters as $C$, and the total number of event mentions as $N$.
\begin{itemize}[leftmargin=*]
    \item \textbf{ARI}~\cite{Hubert1985ComparingPA} measures the similarity between two cluster assignments. Let $TP(TN)$ denote the number of element pairs in the same (different) cluster(s) in both $C^*$ and $C$. Then, ARI is calculated as follows:
    \begin{displaymath}
     \small
        \text{ARI} = \frac{\text{RI} - \mathbb{E}(\text{RI})}{\max \text{RI} - \mathbb{E}(\text{RI})},\quad \text{RI} = \frac{TP+TN}{N},
    \end{displaymath}
    where $\mathbb{E}(\text{RI})$ is the expected RI of random assignments.
    
    \item \textbf{NMI} denotes the normalized mutual information between two cluster assignments and is widely used in previous studies. Let $\text{MI}(\cdot;\cdot)$ be the Mutual Information between two cluster assignments, and $\text{H}(\cdot)$ denote the Entropy. Then the NMI is formulated as follows:
    \begin{displaymath}
     \small
        \text{NMI} = \frac{2 \times \text{MI}(C^*; C)}{\text{H}(C^*)+\text{H}(C)}.
    \end{displaymath}
    
    \item \textbf{BCubed}~\cite{Bagga1998EntityCC} estimates the quality of the generated cluster assignment by aggregating the precision and recall of each element. B-Cubed precision, recall, and F1 are thus calculated as follows:
    \begin{equation}
	{\small
	\begin{split}
        \text{BCubed-P} = \frac{1}{N}\sum_{i=0}^N \frac{|C(e_i) \cap C^*(e_i)|}{|C(e_i)|} \nonumber \\
        \text{BCubed-R} = \frac{1}{N}\sum_{i=0}^N \frac{|C(e_i) \cap C^*(e_i)|}{|C^*(e_i)|} \nonumber \\
        \text{BCubed-F1} = \frac{2}{\text{BCubed-P}^{-1}+\text{BCubed-R}^{-1}} \nonumber 
	\end{split}
	}
	\end{equation}
    where $C^*(\cdot)$ ($C(\cdot)$) is the mapping function from an element to its ground truth (predicted) cluster.
    
    \item \textbf{ACC} measures the quality of the clustering results by finding the permutation function from predicted cluster IDs to ground truth IDs that gives the highest accuracy. Let $y_i$ ($y^*_i$) denote the $i$-th element's predicted (ground truth) cluster ID, the ACC is formulated as follows:
    
    \begin{displaymath}
        \small \text{ACC} = \max_{\sigma \in Perm(k)} \frac{1}{N} \sum_{i=1}^N \mathbbm{1}(y^*_i = \sigma(y_i))
    \end{displaymath}
    where $k$ is the number of clusters for both $C^*$ and $C$, $Perm(k)$ is the set of all permutation functions on the set $\{1, 2, \dots, k\}$, and $\mathbbm{1}(\cdot)$ is the indicator function. 
    
\end{itemize}

\section{Pandemic Dataset Creation}

We follow a similar approach in~\cite{Li2021DocumentLevelEA} to construct our Pandemic Dataset.
First, we resort to Wikipedia lists to get a set of Wikipedia articles related to disease outbreaks\footnote{\small Specifically, we use the list \url{https://en.wikipedia.org/wiki/List_of_epidemics}}.
Then, we extract the news article links from the ``references'' section of those Wikipedia article pages.
Finally, we crawl these news articles based on their above extracted links\footnote{\small We use the crawler tool at \url{https://github.com/codelucas/newspaper}.} and construct a corpus related to disease outbreaks.

\section{Intrusion Test Construction}

Given the top-5 tuples of each detected type, we inject a randomly sampled tuple from the top results of other types to serve as a negative sample. 
For methods that have cluster centers, we rank tuples within each cluster by their distances to the center. Otherwise, we rank tuples according to their frequencies in the corpus.
Then, the intrusion questions from all compared methods are randomly shuffled to avoid bias. 
Three annotators\footnote{\small All three annotators are not in the author list of this paper and provide independent judgements of the tuple quality.} are asked to identify the injected tuples independently, and we take the average of their labeling accuracy to show the quality of the generated event types.

\section {More \ours Outputs}

Table~\ref{table:benchmark_cases} and Table~\ref{table:pandemic_cases_suppl} list example outputs of \ours on ACE/ERE and Pandemic datasets, respectively.

\begin{table*}[t]
\small
    \centering
    \scalebox{.79}{
    \begin{tabular}{m{2.0cm} m{3.5cm}m{13.4cm}}
        \toprule
         \textbf{Event Type} & \textbf{Top Ranked \pair{Predicate Sense, Object Head} Pairs} & \quad \quad    \textbf{Example Sentences in Corpus}\\
         
        \midrule
        \vspace{0.3cm}
         \vbox{\nop{\hbox{\strut Justice:}}\hbox{\strut Arrest-Jail}\nop{\hbox{\strut (ACE)}}}  & \begin{itemize}[label={},leftmargin=-1px]
        \setlength\itemsep{-.5em}
                \vspace{-.3cm}
            \item \pair{arrest\_0, protester}
            \item \pair{arrest\_0, militant}
            \item \pair{arrest\_0, suspect}
            \vspace{-.3cm}
        \end{itemize} &  
        \vspace{-.3cm}
        \begin{itemize}
        \setlength\itemsep{-.5em}
            \item For the most part the marches went off peacefully, but in New York a small group of \emph{\underline{protesters}} were \textbf{arrested} after they refused to go home at the end of their rally, police sources said.
            \item On Tuesday, Saudi security officials said three suspected al-Qaida \emph{\underline{militants}} were \textbf{arrested} in Jiddah, Saudi Arabia, in sweeps following the near-simultaneous suicide attacks on three residential compounds on the outskirts of Riyadh on May 12.
            \item can owe tell us exactly the details, the precise details of how you \textbf{arrested} the \emph{\underline{suspect}}?
            \vspace{-.3cm}
        \end{itemize}
\\
        \midrule
        \vspace{0.3cm}
         \vbox{\hbox{\strut Build$^\nabla$}\nop{\hbox{\strut (ACE)}}}  & \begin{itemize}[label={},leftmargin=-1px]
        \setlength\itemsep{-.5em}
                \vspace{-.3cm}
            \item \pair{build\_0, facility}
            \item \pair{build\_0, center}
            \item \pair{build\_0, housing}
            \vspace{-.3cm}
        \end{itemize} &  
        \vspace{-.3cm}
        \begin{itemize}
        \setlength\itemsep{-.5em}
            \item Plans were underway to \textbf{build} destruction \emph{\underline{facilities}} at all other locations but now the Bush junta has removed from its proposed defense budget for fiscal year 2006 all but the minimum funding for these destruction projects.
            \item Virginia is apparently going to be \textbf{build} a data \emph{\underline{center}} in Richmond, a back-up data center , and a help desk/call center as a follow-on to the creation of VITA, the Virginia Information Technology Agency.
            \item The Habitat for Humanity might be a good one to consider, since their expertise is in \textbf{building} \emph{\underline{housing}}, which of course is so beadly needed over there at this time.
            \vspace{-.3cm}
        \end{itemize}
\\
        \midrule
        \midrule
        \vspace{0.3cm}
         \vbox{\nop{\hbox{\strut Transaction:}}\hbox{\strut Transfer-Money}\nop{\hbox{\strut (ERE)}}}  & \begin{itemize}[label={},leftmargin=-1px]
        \setlength\itemsep{-.5em}
                \vspace{-.3cm}
            \item \pair{fund\_0, activity}
            \item \pair{fund\_0, operation}
            \item \pair{fund\_0, people}
            \vspace{-.3cm}
        \end{itemize} &  
        \vspace{-.3cm}
        \begin{itemize}
        \setlength\itemsep{-.5em}
            \item The grants will \textbf{fund} advisory \emph{\underline{activities}}, including local capacity building, infrastructure development, product development, and development of local insurance companies' capacity to provide index-based insurance products.
            \item The White House had hoped to hold off asking for more money to \textbf{fund} military \emph{\underline{operations}} in Iraq and Afghanistan until after the election, but with costs rising faster than expected, it sent a request for an early installment of \$25 billion to Congress this week.
            \item Watch 'Secret Pakistan' on the BBC iPlayer , it's an awesome two part documentary about how Pakistan has been supporting and \textbf{funding} these \emph{\underline{people}} for years.
            \vspace{-.3cm}
        \end{itemize}
\\
        \midrule
        \vspace{0.3cm}
         \vbox{\hbox{\strut Bombing$^\nabla$}\nop{\hbox{\strut (ERE)}}}  & \begin{itemize}[label={},leftmargin=-1px]
        \setlength\itemsep{-.5em}
                \vspace{-.3cm}
            \item \pair{bomb\_0, factory}
            \item \pair{bomb\_0, checkpoint}
            \item \pair{bomb\_0, base}
            \vspace{-.3cm}
        \end{itemize} &  
        \vspace{-.3cm}
        \begin{itemize}
        \setlength\itemsep{-.5em}
            \item He \textbf{bombed} the Aspirin \emph{\underline{factory}} in 1998 (which turned out to have nothing to do with Bin Laden) the week he revealed he had been lying to us for eight months about Lewinsky.
            \item Prosecutors then also pointed to the men’s suicide bomber training in 2011 in Somalia and association with Beledi, who prosecutors said \textbf{bombed} a government \emph{\underline{checkpoint}} in Mogadishu that year.
            \item Once the war breaks out, Iran will immediately use all kinds of missiles to \textbf{bomb} the military \emph{\underline{bases}} of the United States in the Gulf and Israel to pieces.
            \vspace{-.3cm}
        \end{itemize}
\\
        \bottomrule
        \end{tabular}
    }
    \vspace*{-.1cm}
    \caption{Example outputs of \ours discovered event types with their associated sentences in ACE and ERE datasets. The first two types come from ACE and the remaining two are from ERE. The event types with superscript ``$^\nabla$'' originally do not exist in human-labeled schemas and are discovered by \ours framework. \textbf{Predicates} are in bold and \underline{\emph{object heads}} are underlined and in italics.}
    \label{table:benchmark_cases_suppl}
    \vspace*{-.0cm}
\end{table*}

\begin{table*}[t]
\small
    \centering
    \scalebox{.79}{
    \begin{tabular}{m{1.7cm} m{3.5cm}m{13.7cm}}
        \toprule
         \textbf{Event Type} & \textbf{Top Ranked \pair{Predicate Sense, Object Head} Pairs} & \quad \quad    \textbf{Example Sentences in Corpus}\\
        \midrule
        \vspace{0.3cm}
         \vbox{\hbox{\strut Spread}\hbox{\strut Virus}}  & \begin{itemize}[label={},leftmargin=-1px]
        \setlength\itemsep{-.5em}
                \vspace{-.3cm}
            \item \pair{spread\_2, virus}
            \item \pair{spread\_2, disease}
            \item \pair{spread\_2, coronavirus}
            \vspace{-.3cm}
        \end{itemize} &  
        \vspace{-.3cm}
        \begin{itemize}
        \setlength\itemsep{-.5em}
            \item What is the best way to keep from \textbf{spreading} the \emph{\underline{virus}} through coughing or sneezing?
            \item Farmers quickly mobilized to fight the misperceptions that pigs could \textbf{spread} the \emph{\underline{disease}}.
            \item In the UK, Asians have been punched in the face, accused of \textbf{spreading} \emph{\underline{coronavirus}}.
            \vspace{-.3cm}
        \end{itemize}
\\
         \midrule
         \vspace{0.4cm}
         \vbox{\hbox{\strut Wear}\hbox{\strut Mask}}  & \begin{itemize}[label={},leftmargin=-1px]
        \setlength\itemsep{-.5em}
                \vspace{-.3cm}
            \item \pair{wear\_1, mask}
            \item \pair{wear\_1, facemasks}
            \item \pair{wear\_1, cover}
            \vspace{-.3cm}
        \end{itemize} &  
        \vspace{-.3cm}
        \begin{itemize}
        \setlength\itemsep{-.5em}
            \item Pence chose not to \textbf{wear} a face \emph{\underline{mask}} during the tour despite the facility's policy.
            \item It should not be necessary for workers to \textbf{wear} \emph{\underline{facemasks}} routinely when in contact with the public.
            \item The WHO offers a conditional recommendation that health care providers also \textbf{wear} a separate head \emph{\underline{cover}} that protects the head and neck.
            \vspace{-.3cm}
        \end{itemize}
        
\\
	\midrule
	\vspace{0.4cm}
        \vbox{\hbox{\strut Prevent}\hbox{\strut Spread}}  & \begin{itemize}[label={},leftmargin=-1px]
        \setlength\itemsep{-.5em}
                \vspace{-.3cm}
            \item \pair{prevent\_1, spread}
            \item \pair{mitigate\_1, spread}
            \item \pair{mitigate\_1, transmission}
            \vspace{-.3cm}
        \end{itemize} &  
        \vspace{-.3cm}
        \begin{itemize}
        \setlength\itemsep{-.5em}
            \item Infection prevention and control measures are critical to \textbf{prevent} the possible \emph{\underline{spread}} of MERS-CoV in health care facilities .
            \item A vaccine can \textbf{mitigate} \emph{\underline{spread}}, but not fully prevent the virus circulating.
            \item Asymptomatic infection could also potentially be directly harnessed to \textbf{mitigate} \emph{\underline{transmission}}.
            \vspace{-.3cm}
        \end{itemize}
         
\\
	\midrule
	\vspace{0.4cm}
        \vbox{\hbox{\strut Delay}\hbox{\strut Gathering}}  & \begin{itemize}[label={},leftmargin=-1px]
        \setlength\itemsep{-.5em}
                \vspace{-.3cm}
            \item \pair{delay\_1, gathering}
            \item \pair{postpone\_1, gathering}
            \item \pair{suspend\_1, gathering}
            \vspace{-.3cm}
        \end{itemize} &  
        \vspace{-.3cm}
        \begin{itemize}
        \setlength\itemsep{-.5em}
            \item The 2020 edition of the Cannes Film Festival, was left in limbo following an announcement from the festival’s organizers that the \emph{\underline{gathering}} could be \textbf{delayed} until late June or early July.
            \item States with EVD should consider \textbf{postponing} mass \emph{\underline{gatherings}} until EVD transmission is interrupted.
            \item On Thursday, leaders of The Church of Jesus Christ of Latter - day Saints told its 15 million members worldwide all public \emph{\underline{gatherings}} would be \textbf{suspended} until further notice .
            \vspace{-.3cm}
        \end{itemize}         

\\
	\midrule
	\vspace{0.4cm}
        \vbox{\hbox{\strut Provide}\hbox{\strut Testing}}  & \begin{itemize}[label={},leftmargin=-1px]
        \setlength\itemsep{-.5em}
                \vspace{-.3cm}
            \item \pair{provide\_1, testing}
            \item \pair{conduct\_1, testing}
            \item \pair{perform\_1, testing}
            \vspace{-.3cm}
        \end{itemize} &  
        \vspace{-.3cm}
        \begin{itemize}
        \setlength\itemsep{-.5em}
            \item Governments are racing to buy medical equipment as a debate intensifies over \textbf{providing} adequate \emph{\underline{testing}}, when it 's advisable to wear masks, and whether stricter lockdowns should be imposed.
            \item Additional \emph{\underline{testing}} is being \textbf{conducted} to confirm that the family members had H1N1 and to try to verify that the flu was transmitted from human to cat.
            \item Additional laboratories \textbf{perform} antiviral \emph{\underline{testing}} and report their results to CDC .
            \vspace{-.3cm}
        \end{itemize}      	         

\\
        \midrule
        \vspace{0.4cm}
        \vbox{\hbox{\strut Warn}\hbox{\strut Country}}  & \begin{itemize}[label={},leftmargin=-1px]
        \setlength\itemsep{-.5em}
                \vspace{-.3cm}
            \item \pair{warn\_1, country}
            \item \pair{warn\_1, authority}
            \item \pair{warn\_1, government}
            \vspace{-.3cm}
        \end{itemize} &  
        \vspace{-.3cm}
        \begin{itemize}
        \setlength\itemsep{-.5em}
            \item WHO uses six phases of alert to communicate the seriousness of infectious threats and to \textbf{warn} \emph{\underline{countries}} of the need to prepare and respond to outbreaks.
            \item The message showed a photo of a letter, written by the operators of the hospital’s oxygen supply plant, \textbf{warning} the \emph{\underline{authorities}} that the supply was running dangerously low .
            \item WHO staff concluded there was a high risk of further spread, and issued a global alert to \textbf{warn} all member \emph{\underline{governments}} of the existence of a new and highly infectious form of ``atypical pneumonia'' on March 12th .
            \vspace{-.3cm}
        \end{itemize}

\\
	\midrule
	\vspace{0.5cm}
        \vbox{\hbox{\strut Vaccinate}\hbox{\strut People}}  & \begin{itemize}[label={},leftmargin=-1px]
        \setlength\itemsep{-.5em}
                \vspace{-.3cm}
            \item \pair{vaccinate\_0, person}
            \item \pair{immunize\_0, people}
            \item \pair{vaccinate\_0, family}
            \vspace{-.3cm}
        \end{itemize} &  
        \vspace{-.3cm}
        \begin{itemize}
        \setlength\itemsep{-.5em}
            \item All \emph{\underline{persons}} in a recommended vaccination target group should be \textbf{vaccinated} with the 2009 H1N1 monovalent vaccine and the seasonal influenza vaccine.
            \item U.K. Will Start \textbf{Immunizing} \emph{\underline{People}} Against COVID-19 On Tuesday, Officials Say.
            \item ``In the Samoan language there is no word for bacteria or virus'' says Henrietta Aviga, a nurse travelling around villages to \textbf{vaccinate} and educate \emph{\underline{families}}.
            \vspace{-.3cm}
        \end{itemize}

         \\
        \bottomrule
        \end{tabular}
    }
    \vspace*{-.1cm}
    \caption{More example outputs of \ours discovered event types with their associated sentences in the corpus. \textbf{Predicates} are in bold and \underline{\emph{object heads}} are underlined and in italics.}
    \label{table:pandemic_cases_suppl}
    \vspace*{-.0cm}
\end{table*}

\end{document}